\documentclass[10pt,twocolumn,letterpaper]{article}

\usepackage{iccv}
\usepackage{times}
\usepackage{epsfig}
\usepackage{graphicx}
\usepackage{amsmath}
\usepackage{amssymb}

\usepackage{tikz}
\usetikzlibrary{fit, shapes, arrows, positioning, calc}
\usepackage{forest}
\usepackage{color}

\usepackage[pagebackref=true,breaklinks=true,letterpaper=true,colorlinks,bookmarks=false]{hyperref}

\iccvfinalcopy 

\def\iccvPaperID{****} 

\begin{document}

\title{DeepArchitect: Automatically Designing and Training Deep Architectures}

\author{Renato Negrinho\\
Carnegie Mellon University\\
{\tt\small negrinho@cs.cmu.edu}
\and
Geoff Gordon\\
Carnegie Mellon University\\
{\tt\small ggordon@cs.cmu.edu}
}

\maketitle


\begin{abstract}
In deep learning, performance is strongly affected by the choice of 
architecture and hyperparameters. 
While there has been extensive work on 
automatic hyperparameter optimization 
for simple spaces, 
complex spaces such as the space of 
deep architectures remain largely unexplored. 
As a result, the choice of architecture is done manually by the human expert 
through a slow trial and error process guided mainly by intuition.
In this paper we describe 
a framework for automatically 
designing and training deep models.
We propose
an extensible and modular language 
that allows the human expert
to compactly represent complex search spaces 
over architectures and their hyperparameters.
The resulting search spaces are 
tree-structured and therefore easy to traverse.
Models can be automatically compiled to computational graphs
once values for all hyperparameters have been chosen. 
We can leverage the structure of the search space 
to introduce different model search algorithms,
such as 
random search, Monte Carlo tree search (MCTS), and 
sequential model-based optimization (SMBO).
We present experiments comparing the different 
algorithms on CIFAR-10 and 
show that MCTS and SMBO outperform 
random search. 
In addition, these experiments 
show that our framework 
can be used effectively
for model discovery,
as
it is possible 
to describe expressive search spaces 
and discover competitive models 
without much effort from the human expert. 
Code for our framework and experiments 
has been made publicly available.
\end{abstract}



\section{Introduction}
\label{sec:intro}
Deep learning has seen a surge in popularity due to
breakthroughs in applications such as 
computer vision, 
natural language processing, and
reinforcement learning~\cite{he2016deep,karpathy2015deep, silver2016mastering,sutskever2014sequence}.
An important observation in much of the recent work is that complex 
architectures are important for achieving high performance~\cite{he2016deep, mnih2013playing}.
Larger datasets and more powerful computing infrastructures are likely to 
increase our ability to effectively train larger, deeper, and more complex architectures.
However, improving the performance of a neural network is not as simple as 
adding more layers or parameters---it 
often requires clever ideas such as creating more
branches~\cite{szegedy2015going} or adding skip connections~\cite{he2016deep}.
Even popular techniques such as dropout~\cite{srivastava2014dropout} and
batch normalization~\cite{ioffe2015batch} 
do not always lead to better performance, 
and need to be judiciously applied to be helpful.

Currently, choosing appropriate values for these architectural hyperparameters
requires close supervision by a human expert, in a trial and error manual search 
process largely guided by intuition.
The expert is burdened by having to make the 
large number of choices 
involved in the specification of a deep model.
Choices interact in non-obvious ways and 
strongly impact performance.
The typical workflow has the expert specify a single model, 
train it, and 
compute a validation score. 
Based on 
the validation score, 
previous experience, and 
information gathered during training, 
the expert decides if the trained model 
is satisfactory or not.
If the model is considered unsatisfactory, 
the expert has to think about model variations
that 
may lead to better performance.

From the perspective of the expert, it would be convenient to 
search over architectures automatically, just as
we search over  
simple scalar hyperparameters, 
such as the learning rate and 
the regularization coefficient.
Ideally, the expert would have control
in setting up the search space to incorporate 
inductive biases about the task being solved and
constraints about computational resources.
Prior to this work, achieving this goal 
was hard because
expressing model search spaces 
using
general hyperparameter optimization tools 
requires the human expert to manually distill a  
set of relevant scalar architectural hyperparameters.

The main contributions of our work are 
\begin{enumerate}
    \item a modular, compositional, and extensible language 
        for compactly 
        representing expressive search spaces over models that
        \begin{enumerate} 
            \item gives control to the human expert over 
                what model variations to consider; 
            \item makes it easy to automatically search for 
                performant 
                models in the search space; 
            \item allows models to be directly compiled to 
                computational graphs 
                without the human expert 
                having to write additional code.
        \end{enumerate}
    \item model search algorithms that rely on 
        the tree-structured search spaces induced by our language   
        to systematically and efficiently 
        search for performant models; namely, we
        \begin{enumerate}
        \item show that by using the constructs 
            in our language, even
            random search can be effective;
        \item compare different  
            model search algorithms experimentally,
            and show that random search is outperformed 
            by 
            algorithms that leverage the 
            structure of the search space to
            generalize more effectively across different models.
        \end{enumerate}
\end{enumerate}

The main differences between our work and previous work are that we 
develop a modular, composable and extensible language, 
focusing on the problem of searching over deep architectures. 
This focus allows the expert to compactly set up a search space, 
search over it, 
and automatically compile models 
to their corresponding computational graphs.
Our language can be seen as an effort to combine the functionalities of
a deep model specification language 
(e.g., Tensorflow~\cite{abadi2016tensorflow}) and 
a structured hyperparameter search language 
(e.g., Hyperopt~\cite{bergstra13modelsearch}).





\section{Related Work}
\label{sec:related_work}


Model search has a long and rich history in machine learning and statistics. 
There has been a wide variety of theoretical and empirical research
in this area%
~\cite{agarwal2011oracle,bergstra11hyperparameter,bergstra12randomSearch,sabharwal2015selecting},
including Bayesian optimization methods~\cite{hutter2011sequential,kandasamy15addBO,snoek12practicalBO}.
However, conventional methods are primarily designed for searching over
hyperparameters living in Euclidean space.
Such methods are ill suited in today's context, 
where 
the discrete architectural choices are 
just as important
as the numerical values of the hyperparameters.
Searching over architectures using current hyperparameter optimization 
algorithms requires 
the expert to distill structural choices into 
scalar hyperparameters.
As a result, typically only a few simple global 
structural hyperparameters are considered,
e.g., the depth of the network or whether to use dropout or not.
This constrains the richness of the search space, 
preventing the expert from finding 
unexpected model variations 
leading to better performance; e.g., 
perhaps dropout is useful only after certain types of layers, or 
batch normalization only helps in the first half of the network. 

Architecture search 
has also been considered under 
the topic of neuroevolution~\cite{stanley2002evolving}, 
which uses evolutionary (i.e., genetic) strategies 
to define and search a space of models.
In classical approaches, neuroevolution attempts to 
jointly choose 
the topology and the parameters of the architecture  
using genetic algorithms. 

Architecture search has received renewed interest recently.
Wierstra \etal
\cite{wierstra2005modeling}, 
Floreano \etal \cite{floreano2008neuroevolution},
and Real \etal \cite{large2017real} 
use evolutionary
algorithms which start from an initial model and 
evolve it based on its validation performance.
Zoph and Le \cite{zoph2016neural} 
propose a reinforcement learning procedure 
based on policy gradient 
for searching for 
convolutional and 
LSTM architectures.
Baker \etal \cite{baker2016designing} 
propose a reinforcement learning procedure 
based on Q-learning 
for searching for 
convolutional architectures. 

Unfortunately all these approaches consider 
fixed hard-coded model search spaces
that do not easily allow the 
human expert to incorporate inductive biases 
about the task being solved, making them unsuitable 
as general tools for architecture search.
For example, evolutionary approaches require 
an encoding for the models in the search space
and genetic operators (e.g., mutation and crossover)
which generate encodings for new models out of 
encodings of old ones.
These aspects are handcrafted and hard-coded
so it is hard for the human expert to 
change the search space in flexible ways.
Perhaps 
different model encodings
or genetic operators can be considered, 
but these knobs give somewhat loose and indirect control over the 
model search space.
The reinforcement learning approaches considered 
suffer from similar issues---the 
search spaces are hard-coded and not easily modifiable. 
None of these approaches have
the compositionality, modularity, and extensibility properties 
of our language.

Bergstra \etal \cite{bergstra11hyperparameter} propose
Tree of Parzen Estimators (TPE), which 
can be used to search over structured hyperparameter spaces, 
and use it to tune the hyperparameters of a Deep Boltzmann Machine.
Bergstra \etal \cite{bergstra13modelsearch} use
TPE to search for values of the hyperparameters 
of a computer vision system, 
and show that it can find better values 
than the best ones previously known.

TPE 
is a general 
hyperparameter search algorithm, 
and therefore requires considerable effort to use---for 
any fixed model search space, using TPE requires 
the human expert to distill the hyperparameters 
of the search space, 
express the search space in Hyperopt~\cite{bergstra13modelsearch}
(an implementation of TPE), and write the code describing 
how values of the hyperparameters in the search space 
compile to a computational graph.
In contrast, our language is modular and composable in 
the sense that: 
\begin{enumerate}
    \item search spaces (defined through modules) 
are constructed compositionally out of 
simpler search spaces (i.e., simpler modules); 
    \item hyperparameters for 
    composite modules are derived automatically 
    from the hyperparameters of simpler modules; 
    \item once values for all hyperparameters of a module have been chosen, 
    the resulting model can be automatically mapped 
    to a computational 
    graph without the human expert having to 
    write additional code.
\end{enumerate}

\section{Roadmap to the DeepArchitect Framework}
\label{sec:roadmap}



Our framework
reduces the problem of searching over models into 
three modular components:
the \emph{model search space specification language}, 
the \emph{model search algorithm}, 
and the \emph{model evaluation algorithm}.

\paragraph{Model Search Specification Language:}
The model search space specification language is built around the concept of a 
modular \emph{computational module}.
This is akin to the concept of a module~\cite{bottou1991framework}
used in deep learning frameworks such as Torch~\cite{torch}:
by implementing the module interface,  
the internal implementation becomes irrelevant.
These modules allow one to 
express easily complex design choices such as 
whether to include a module or not, 
choose between modules of different types, 
or choose how many times to repeat a module structure.
The main insight is that 
complex modules can be created \emph{compositionally} out of simpler ones.
The behavior of complex modules
is generated automatically out of the behavior of
simpler modules. 
Furthermore, our language is extensible, 
allowing the implementation of new types of modules 
by implementing a high-level interface 
local to the module.

\paragraph{Model Search Algorithm:}
The way the model search space is explored is determined by 
the \emph{model search algorithm}. 
This part of the framework decides 
how much effort to allocate to each part 
of the search space based on 
the performance observed for previous models.
The model search algorithm typically requires a
\emph{model evaluation algorithm} that 
computes the performance of a fully specified model.
The search algorithm will then use this information to determine 
which models to try next.
The search algorithm interacts with the search space only through a 
minimal interface that allows it to traverse the space of models
and evaluate models discovered this way.
This interface is the same irrespective of the specific 
search space under consideration.
We experiment with 
different search algorithms,
such as
Monte Carlo tree search~\cite{browne2012survey} and 
Sequential Model Based Optimization~\cite{hutter2011sequential}.

\paragraph{Model Evaluation Algorithm:}
Having fully specified a model, i.e., having reached a leaf in the 
tree defined by our model search space, we can evaluate how good this model is  
according to some criterion defined by the expert.
This typically involves training the model on a training set and 
evaluating it on a validation set.
The training procedure often has multiple hyperparameters that can 
be tuned
(e.g., the choice of the 
optimization algorithm and 
its hyperparameters, 
and the learning rate schedule).
If the expert does not know how to 
write down a reasonable training procedure 
for every model in the search space,
the expert can introduce hyperparameters for the evaluation algorithm 
and search over them  
using our specification language.



\paragraph{}
Any of the above components 
can be changed, improved, or extended,
while keeping the others fixed. 
The fact that different components interact only through 
well-defined interfaces makes it possible to 
extend and reuse this framework. 
We believe that DeepArchitect will be an interesting platform 
for future research in deep learning and 
hyperparameter tuning 
for architecture search.

\section{Model Search Space Specification Language}
\label{sec:model_space_spec_lang}
\subsection{Search Space Definition}
\label{ssec:space_spec}

The \emph{computational module} is the fundamental unit 
of our model search space specification language.
We define a computational module as a function 
\begin{equation}
f : n \to 
    \left( \mathcal H \to ( \mathbb R ^p  \to (\mathbb R ^{n} \to \mathbb R ^{m} ) )
    \right),
\label{eq:module_def}
\end{equation}
where
$n$ is the dimensionality of the \emph{input},
$\mathcal H$ is the set of valid values for the \emph{hyperparameters},
$p$ is the number of \emph{parameters}, and 
$m$ is the dimensionality of the \emph{output}.
The set
$\mathcal H$ can be structured
or simply the cross product of scalar hyperparameter sets,
i.e., $\mathcal H = \mathcal H _1 \times \ldots \times \mathcal H _H$,
where $H$ is the number of scalar hyperparameters.
The set $\mathcal H$ is assumed to be discrete in both cases. 


Definition~\eqref{eq:module_def} merits some discussion. 
For conciseness 
we have not explicitly represented it, but 
\emph{the number of parameters $p$ and 
the output dimensionality $m$
can both be
functions of  
the input dimensionality $n$ 
and the chosen hyperparameter values $h \in \mathcal H$}.
For example, an affine module
with
$h$ dense hidden units
has output dimensionality $m = h$
and number of parameters 
$p = (n + 1) h$:  
 a 
weight matrix $W \in \mathbb R ^{h \times n}$
and a bias vector $b \in \mathbb R ^h$.
A similar reasoning can be carried out 
for a convolutional module:
the number of parameters
$p$ depends on 
the input dimensionality, 
the number of filters, 
and the size of the filters;
the dimensionality of the output $m$ depends on 
the input dimensionality,
the number of filters, 
the size of the filters,
the stride, 
and the padding scheme.
The fact that $p$ and $m$ are functions of
the input dimensionality
and the chosen hyperparameter values
is one of the main observations  
that allows us to do architecture search---\emph{once we 
know the input dimensionality and
have fixed values for the hyperparameters,
the structure of the computation 
performed by the module is determined},  
and this information can be propagated to other modules.
We say that a module is \emph{fully specified}
when values for all hyperparameters of the module 
have been chosen and the input dimensionality is known. 

We focus
on search spaces for architectures that have 
a \emph{single input terminal} and 
a \emph{single output terminal}.
By this, we only mean that the input and output of the 
module have to be a \emph{single} tensor of arbitrary order and dimensionality.
For example, convolutional modules take as input 
an order three tensor and 
return as output 
an order three tensor,
therefore they are
single-input single-output modules
under our definition.
We also assume that the output of a module 
is used as input to at most a single module, i.e., 
we \emph{assume no output sharing}.

These restrictions were introduced to 
simplify exposition. 
The single-input single-output case
with no sharing is simpler to develop 
and exemplifies the main ideas that 
allow us to develop a framework for automatic architecture search.
The ideas developed in this work extend naturally 
to the multiple-input multiple-output case with sharing.
Additionally, often 
we can represent modules that 
are not single-input single-output
by defining 
new modules that encapsulate many signal paths 
from input to output.
For example, a residual module~\cite{he2016deep}
can be treated in our framework 
by noting that it is single-input before 
the skip connection split
and single-output after 
the skip connection merge.
Many top performing architectures,
such as 
AlexNet~\cite{krizhevsky2012imagenet}, 
VGG~\cite{simonyan2014very}, and 
ResNet~\cite{he2016deep},
are captured in our language.


We distinguish between 
\emph{basic computational modules} and
\emph{composite computational modules}.
Basic modules 
do some well defined transformation. 
Affine, batch normalization, and dropout are examples of basic modules.
Composite modules are defined in terms of other (composite or basic) modules,
i.e., the instantiation of a composite module 
takes other modules as arguments. 
Composite modules may introduce 
hyperparameters of their own 
and inherit hyperparameters of 
the modules taken as arguments.
For example, an \texttt{Or} module 
takes a list of modules and 
chooses one of the modules
to use.
It
introduces a discrete hyperparameter 
for 
which module to use, 
and chooses values for the 
hyperparameters of the chosen module;
the hyperparameters available are conditional on
the choice of the module to use.
Most of the representational power of our language 
arises from the compositionality 
of composite and basic modules.

The ideas developed in this section are perhaps 
best illustrated with an example.
See Figure~\ref{fig:ss_program_ex} for 
the definition of an example search space
in LISP-like pseudocode that closely parallels our implementation.
The search space,
which results from the composition of several modules,
and therefore is also a module itself, 
encodes $24$ different models, corresponding 
to the different $24$ possible paths 
from the root to the leaves of the tree.
The space is defined using three composite modules 
(\texttt{Concat}, \texttt{MaybeSwap}, and \texttt{Optional})
and five basic modules 
(\texttt{Conv2D}, \texttt{BatchNormalization}, \texttt{ReLU}, \texttt{Dropout},
and \texttt{Affine}).
\texttt{Concat} introduces no additional hyperparameters, 
but it has to specify all the modules that have been delegated to it;
\texttt{MaybeSwap} introduces a binary hyperparameter that encodes 
whether to swap the order of the pair of modules or not;
\texttt{Optional} introduces a binary hyperparameter that encodes 
whether to include the module or not.
The behavior of the basic modules in Figure~\ref{fig:ss_program_ex} 
is relatively straightforward:
\texttt{Conv2D} takes lists of possible values for
the number of filters,
the size of the filters, and 
the stride;
\texttt{BatchNormalization} and \texttt{ReLU} 
have no hyperparameters;
\texttt{Dropout} takes a list for the possible values for 
the dropout probability;
\texttt{Affine} takes a list for the possible values of 
the number of hidden units.
\begin{figure}[t]
{\raggedleft 
\begin{verbatim}
(Concat 
    (Conv2D [32, 64] [3, 5] [1])
    (MaybeSwap BatchNormalization ReLU)
    (Optional (Dropout [0.5, 0.9]))
    (Affine [10]))
\end{verbatim} 
}
\caption{A simple search space with $24$ different models. 
    See Figure~\ref{fig:ss_tree_ex} for
    a path through the search space.}
\label{fig:ss_program_ex}
\end{figure}

Choosing different values for the hyperparameters 
of the composite modules may
affect the structure of the resulting architecture,
while choosing different values for the hyperparameters
of the basic modules 
only affects the structure of the corresponding local transformations.
The search space of Figure~\ref{fig:ss_program_ex} results 
from the composition of basic and composite modules;
therefore it is a module itself and 
can be characterized by its 
input, 
output, 
parameters, 
and hyperparameters.
Our set of composite modules in not minimal: 
e.g., 
given an 
\texttt{Empty} basic module, 
which 
has no hyperparameters or parameters and 
simply does the identity transformation,  
and a 
\texttt{Or} composite module,
which introduces an extra hyperparameter
encoding the choice of a specific module in its list, 
the composite modules 
\texttt{Optional} and \texttt{MaybeSwap} can be defined
as
\texttt{(Optional B) = (Or Empty B)}
and 
\texttt{(MaybeSwap B1 B2) = (Or (Concat B1 B2), (Concat B2 B1))}.

\subsection{Search Space Traversal}
\label{ssec:traversal}

Given a search space defined by a module, 
there is an underlying tree over fully specified models:
we build this tree by 
sequentially 
assigning values to each of the 
hyperparameters of the module.
Each internal node in the tree corresponds to some partial 
assignment to the hyperparameters of the module,
and each terminal node (i.e., each leaf) 
corresponds to a fully specified model.
We can also think about an internal node as corresponding to 
the state of a module 
before assigning a value to the next unassigned hyperparameter. 
The branching factor of a node corresponds to the number 
of possible values for the hyperparameter under consideration
at that node, 
and traversing a specific edge from that node to a child corresponds 
to assigning the value encoded by that edge to the 
hyperparameter under consideration.
As a tree has a single path between the root and any leaf, 
the paths from root to leaves are in one-to-one correspondence 
with fully specified models.
A leaf is reached when there are no hyperparameters left to specify. 

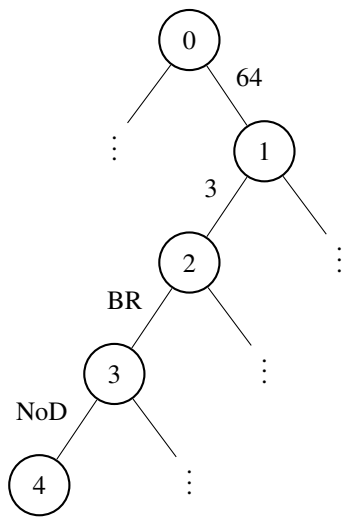
\begin{figure}[t]
\centering
\begin{forest}
for tree={l sep=15pt, s sep=40pt, thick}
[0,circle,draw, minimum size = 8mm, 
    [ $\vdots$ ]
    [1,circle, draw, minimum size = 8mm, 
        edge label={ node[midway,above right] {64} } 
        [2, circle, draw, minimum size = 8mm, 
            edge label={ node[midway,above left] {3} } 
            [3,circle, draw, minimum size = 8mm,
                edge label={ node[midway,above left] {BR} }  
                [4,circle, draw, minimum size = 8mm,
                    edge label={ node[midway,above left] {NoD} } ]
                [ $\vdots$ ]    
            ]
            [ $\vdots$ ]
        ]
        [ $\vdots$ ]
    ]
]
\end{forest}
\caption{A path through the 
 search space
of Figure~\ref{fig:ss_program_ex}.
The model encoded by this path has a 
convolutional module with 
$64$ filters of
size $3\times 3$, with stride $1$, 
followed by batch normalization, ReLU and affine modules.
The model does not use dropout. 
Branches encoding hyperparameters with a single choice were omitted. 
}
\label{fig:ss_tree_ex}
\end{figure}

In Figure~\ref{fig:ss_tree_ex} we have drawn a path through the 
search space of  
Figure~\ref{fig:ss_program_ex}
from the 
root (labeled node $0$), 
where all hyperparameters are unassigned,
to a terminal node (labeled node $4$), 
where all hyperparameters have been assigned values.
Each branch in the tree corresponds to the assignment of some 
value to some hyperparameter.
At node $0$, 
we are choosing between 
$32$ or $64$ filters;
at node $1$, 
we are choosing between 
filters of size $3$ or $5$;
at node $2$, we are choosing between  
applying batch normalization 
before or after ReLU;
at node $3$, we are choosing 
whether to do dropout or not.
Node $4$ is terminal and corresponds to 
a fully specified model.
Decisions at each node are conditional on 
decisions previously made.
Internal nodes with a single child
(i.e., branches for hyperparameters with a single possible value)
have been collapsed and omitted from Figure~\ref{fig:ss_program_ex}.
Other paths may have different lengths, e.g., 
picking a path through the right child of node $3$ 
corresponds to adding a \texttt{Dropout} module, 
which requires an additional hyperparameter choice for 
the dropout probability
when compared to the path from the root to node $4$. 

Search spaces arising from module composition 
have their traversal functionality 
automatically derived from the traversal 
functionality of their component modules:
a basic module knows how to 
sequentially assign values to its hyperparameters,
and a composite module knows how to 
sequentially assign values to its hyperparameters and  
call the sequential assignment functionality for its component modules.
This behavior is akin to recursive expression evaluation in programming languages.

To traverse the search space, 
i.e., to assign values to all hyperparameters 
of the module defining the 
search space, 
all that it is needed is that each module knows how to 
sequentially specify itself.
\emph{Modules resulting from the composition of modules will then be 
automatically sequentially specifiable}.
The three \emph{local} operations that a module 
needs to implement for traversal are:
\emph{to test whether it is fully specified}
(i.e., whether it has reached a leaf yet);
if it is not specified, 
\emph{to return which hyperparameter it is specifying and what are the 
possible values for it};
and given a choice for 
the current hyperparameter under consideration, 
\emph{to traverse the edge to the child of the current node 
corresponding to chosen value}.

\subsection{Compilation}
\label{ssec:compilation}

Once values for 
all hyperparameters of a module have been chosen, 
the 
fully specified model
can be
automatically 
mapped to its 
corresponding
\emph{computational graph}. 
We call this mapping
\emph{compilation}.
This operation only requires that each module knows how to locally map 
itself to a computational graph: compilation is derived recursively 
from the compilation of simpler modules.
For example, if we know how to compile 
\texttt{Conv2D}, \texttt{ReLU}, and \texttt{Or} modules, 
we will automatically be able to compile
all modules built from them.
This behavior is also similar 
to recursive expression evaluation in 
programming languages.

\section{Model Search Algorithms}
\label{sec:model_search}

In this section, we consider different search algorithms that 
are built on top of the functionality described above.
Some of these algorithms rely on the fact that the search space is 
tree structured.
One of the challenges of our setting is that deep models are expensive to 
train, so unless we have access to extraordinary computational resources, 
only a moderate number of evaluations 
will be practical.










\subsection{Random Search}
\label{ssec:rand_search}

Random search is the simplest algorithm that we can consider. 
At each node of the tree, we choose 
an outgoing edge uniformly at random,  
until we reach a leaf node (i.e., a model).
Even just random search is interesting,
as 
the model search space specification language allows us to 
capture expressive structural search spaces. 
Without our language, 
randomly selecting an interesting architecture to 
try would not be possible
without considerable effort from the human expert.

\subsection{Monte Carlo Tree Search}
\label{ssec:mcts_search}

Monte Carlo tree search (MCTS)~\cite{browne2012survey, kocsis2006bandit}
is an approximate planning technique that has been used 
effectively in many domains~\cite{silver2016mastering}.
Contrary to random search, MCTS uses the information gathered so far
to steer its policy towards better performing parts of the 
search space.
MCTS maintains a search tree that is  
expanded incrementally one node at a time.
MCTS uses two policies: 
a \emph{tree policy}, 
which determines 
the path to be traversed from the root 
to the frontier
of the already expanded tree; and
a \emph{rollout policy}, 
which determines
the path to be traversed from the frontier
of the already expanded tree 
until a leaf is reached.
Once a leaf is reached, the model encoded by it is 
evaluated 
(e.g., trained on the training set and 
evaluated on the validation set), 
and the resulting score is used to update the statistics of the 
nodes 
in the currently expanded tree 
in the path to the leaf.
Each node in the expanded tree keeps statistics 
about the number of times it was visited and the
average score of the models that were evaluated in the 
subtree at that node.
The rollout policy is often simple, e.g., 
the random policy described in Section~\ref{ssec:rand_search}.

The tree policy typically uses an
upper confidence bound (UCB)
approach. 
Let $n$ be the number of visits of a node $v \in \mathcal T$,
where $\mathcal T$ denotes the currently expanded tree, and 
$n _1, \ldots, n _b$ and 
$\bar X _1, \ldots, \bar X  _b$ be, respectively, 
the number of visits and 
the average scores of the $b$ children of $v$.
The tree policy at $x$ chooses to traverse 
an edge corresponding to a child maximizing the UCB score:
\begin{align}
    \max _{i \in \{1, \ldots, b \} } 
        \bar X _i + 
            2 c 
                \sqrt
                    { \frac {2 \log n } 
                    { n _i }},
\end{align}
where $c \in \mathbb R _+$ is a 
constant capturing the trade-off between 
exploration and exploitation---larger 
values of $c$ correspond to 
larger amounts of exploration.
If at node $x$, some of its children 
have not been added to the tree, there will be some 
$i \in \{1, \ldots, b \} $ for which $n _i = 0$;
in this case we define the UCB score to be infinite, 
and therefore, unexpanded children
always take precedence over 
expanded children.
If multiple unexpanded children are available, 
we expand one uniformly at random.

\subsection{Monte Carlo Tree Search with Tree Restructuring}
\label{ssec:restruct_mcts_search}

When MCTS visits 
a node in the expanded part of 
the tree, 
it has to expand all children of that node before 
expanding any children of its currently expanded children. 
This is undesirable 
when there are hyperparameters 
that can take a 
large number of related values.

We often consider
hyperparameters which take \emph{numeric} values,
and  
similar values result in similar performance.
For example, choosing between
$64$ or $80$ filters for a convolutional module 
might not have a dramatic impact 
on performance. 

A way of addressing such hyperparameters
is to restructure the branches of the 
tree by doing bisection.
Assume that the set of hyperparameters has 
a natural ordering.
At a node, 
rather than committing directly 
to a value of the 
hyperparameter, 
we commit sequentially---first we decide 
if we are choosing a value in the 
first or second half of the set of 
hyperparameters, 
and then we recurse on the chosen half
until we have narrow it down to a single value.
See an example tree in Figure~\ref{fig:tree_branch} and 
the corresponding 
restructured tree in Figure~\ref{fig:tree_bisect}.
\begin{figure}[thb]
\centering
\begin{forest}
for tree={l sep=16pt, s sep=16pt, thick}
[0,circle,draw, minimum size = 8mm, 
    [1,circle, draw, minimum size = 8mm,
        edge label={ node[midway,left] {16} } ]
    [2,circle, draw, minimum size = 8mm,
        edge label={ node[midway,left] {32} } ]
    [3,circle, draw, minimum size = 8mm,
        edge label={ node[midway,left] {48} } ]
    [4,circle, draw, minimum size = 8mm,
        edge label={ node[midway,right] {64} } ]
    [5,circle, draw, minimum size = 8mm,
        edge label={ node[midway,right] {80} } ]
]
\end{forest}
\caption{A tree encoding an hyperparameter and its five possible values.
MCTS applied to this tree is sample-inefficient 
as there is no sharing 
of information between the different child nodes. 
See also Figure~\ref{fig:tree_bisect}.
}
\label{fig:tree_branch}
\end{figure}
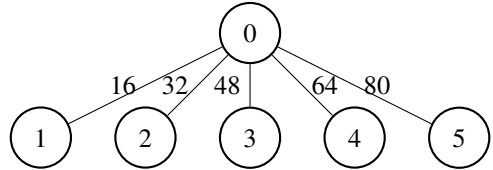

\begin{figure}[thb]
\centering
\begin{forest}
for tree={l sep=16pt, s sep=16pt, thick}
[0,circle,draw, minimum size = 8mm, 
    [,circle, draw, minimum size = 8mm,
            edge label={ node[midway,left] {[16, 32, 48]} }  
        [,circle, draw, minimum size = 8mm,
            edge label={ node[midway,left] {[16, 32]} } 
            [1,circle, draw, minimum size = 8mm,
                edge label={ node[midway,left] {16} } ]  
            [2,circle, draw, minimum size = 8mm,
                edge label={ node[midway,right] {32} } ]    
        ]
        [3,circle, draw, minimum size = 8mm,
            edge label={ node[midway,right] {48} } ]
    ]
    [,circle, draw, minimum size = 8mm,
            edge label={ node[midway,right] {[64, 80]} }  
        [4,circle, draw, minimum size = 8mm,
            edge label={ node[midway,left] {64} } ]
        [5,circle, draw, minimum size = 8mm,
            edge label={ node[midway,right] {80} } ]
    ]
]
\end{forest}
\caption{The result of restructuring the tree 
in Figure~\ref{fig:tree_branch} with bisection.
MCTS applied to this tree results in more sharing when compared to the 
original tree. 
For example, sampling a path reaching node~$1$ provides information about 
nodes~$1$, $2$, and $3$.
}
\label{fig:tree_bisect}
\end{figure}
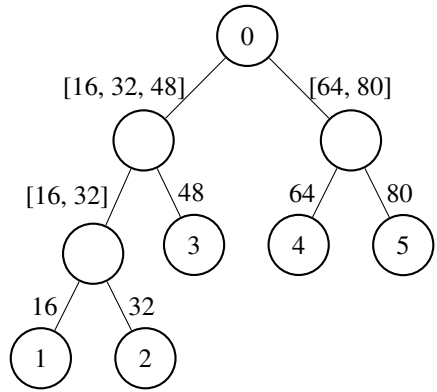

Tree restructuring involves a tradeoff between depth and breadth: 
the tree in Figure~\ref{fig:tree_branch} has depth $1$, 
while the tree in Figure~\ref{fig:tree_bisect} has depth $3$.
The restructured tree can have better properties in the sense that 
there more sharing between different values of the 
hyperparameters.
We could also consider restructured trees with 
branching factors different than two,
again trading off  
depth and breadth.
If the branching factor of the restructured tree 
is larger than the number of children of the hyperparameter, 
the restructuring has no effect, i.e., 
the original and restructured trees are equal.
The restructuring operation allows MCTS 
to \emph{effectively consider
hyperparameters with a large number of possible values}.

\subsection{Sequential Model Based Optimization}
\label{ssec:smbo_search}

MCTS is tabular in the sense that 
it keeps statistics for each node in the tree.
While the restructuring operation described 
in Section~\ref{ssec:restruct_mcts_search}
increases sharing between different hyperparameter 
values, it still suffers from the problem that 
nodes have no way of 
sharing information other than through 
common ancestors.
This is problematic because differences in hyperparameter 
values at the top levels of the tree lead to little sharing between
models, even if the resulting models happen to 
be very similar.

    Sequential Model Based Optimization (SMBO)~\cite{hutter2011sequential} 
allows us to address 
this problem by introducing a surrogate function which can 
be used to capture relationships between models 
and how 
promising it is to evaluate any specific model.
The surrogate function can use expressive 
features to 
capture architecture patterns that influence performance, 
e.g., features about sequences of 
basic modules that occur in the model.

The surrogate function can then be optimized to choose
which model to evaluate next.
    Exactly optimizing the surrogate function over a search space
can be difficult 
as often 
there is a combinatorially large number 
of models. 
To approximately optimize the surrogate function, we 
do some number of random rollouts from the root of the tree 
until we hit leaf nodes (i.e., models), 
we evaluate the surrogate function (i.e., we determine, according to the
surrogate function, how promising it is to evaluate that model), 
and evaluate the model that has the highest score according to the 
surrogate function.
We also introduce an exploratory component
where we flip a biased coin 
and choose between evaluating a random model
or evaluating the
best model according to the surrogate function.
The surrogate function is updated after each evaluation.

In our experiments, we use a simple surrogate function:
we train a ridge regressor to
predict model performance, 
using the models evaluated so far 
and their corresponding performances
as training data.
We only use features based on $n$-grams of sequences of 
basic modules,
disregarding the values of the hyperparameters.
More complex features, surrogate functions, and training losses 
are likely to lead to better search performance, 
but we leave these to future work.


\begin{figure*}[htp]
    \centering
    \includegraphics[width=.32\linewidth]{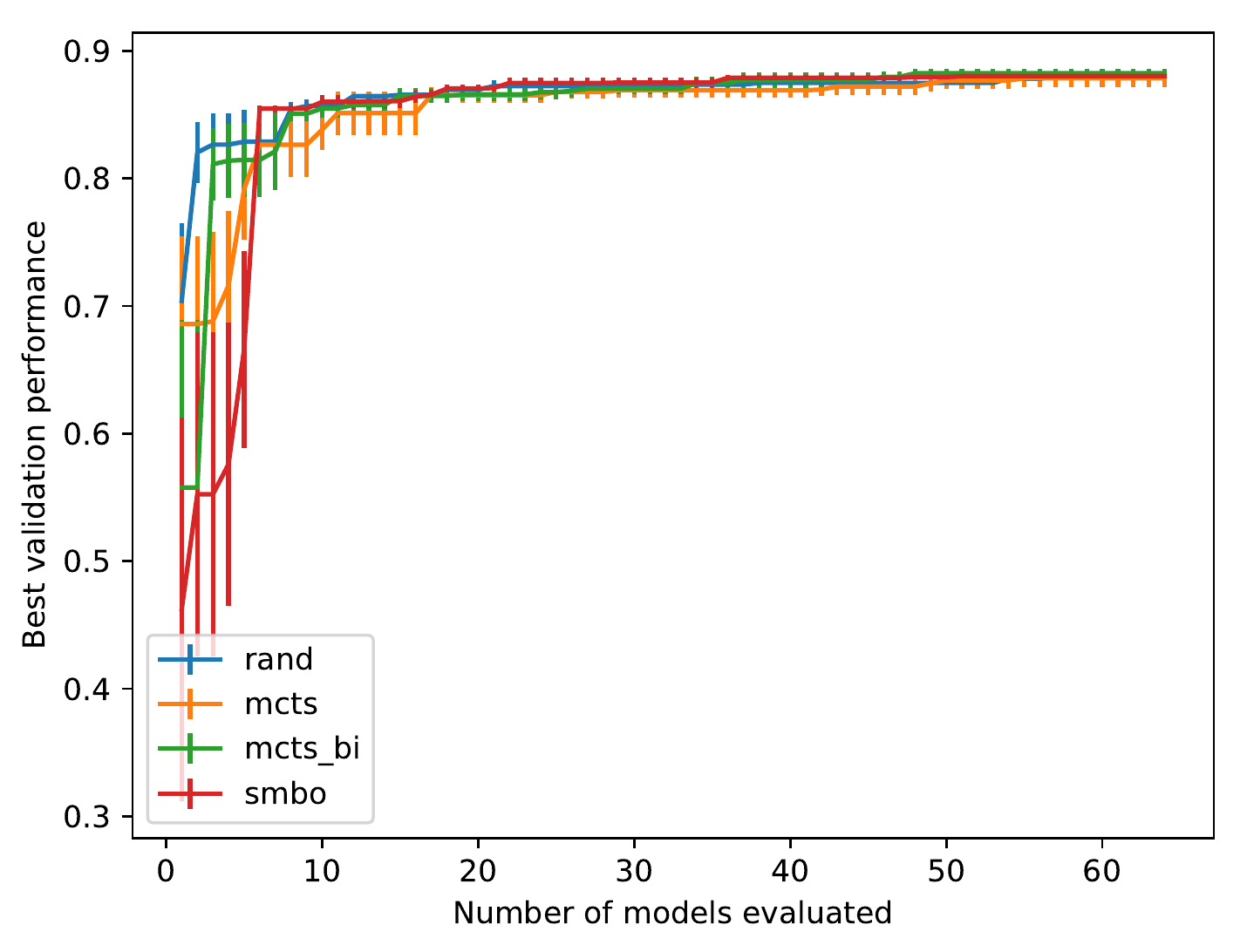}
    \includegraphics[width=0.32\linewidth]{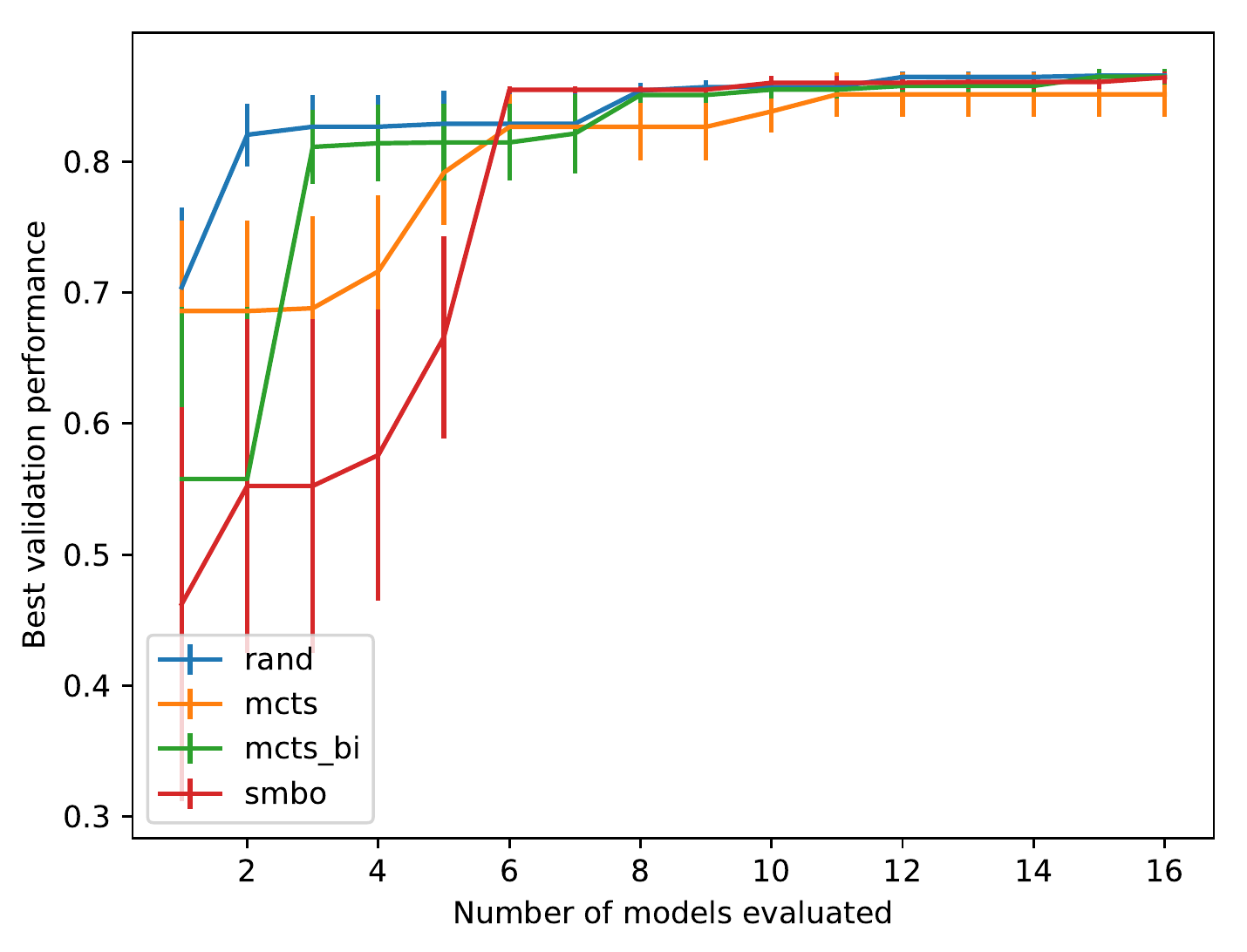}
    \includegraphics[width=0.32\linewidth]{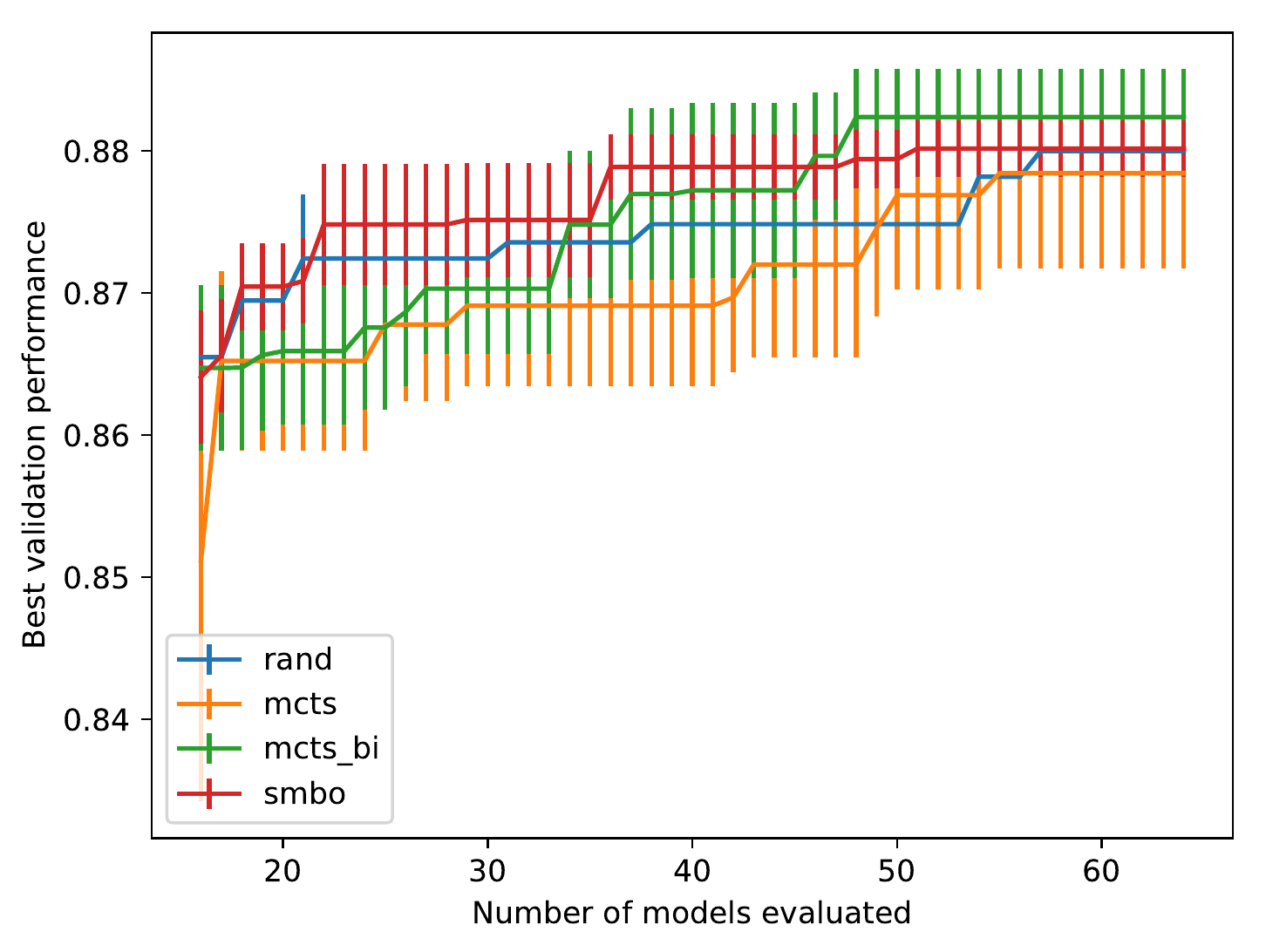}
    \caption{Average maximum validation score achieved as a function of the number of 
        evaluation across five repetitions. The error bars indicate standard error.
        The two plots on the right have the same results as the plot on the left, 
        but are zoomed in for better visualization. }
    \label{fig:searcher_comp}
\end{figure*}


\section{Model Evaluation Algorithms}
\label{sec:model_eval}

As a reminder, once we assign values to all hyperparameters of the module 
defining the search space, 
we need to compute a score for the resulting model, 
i.e., a score for the path from the root to the corresponding leaf
encoding the model to evaluate.
The specific way to 
compute scores
is defined by the human expert,
and it typically amounts to training the model on 
a training set and evaluating the trained model on a validation set.
The score of the model is 
the resulting validation performance.
The training process often has its own hyperparameters, 
such as: 
what optimization algorithm to use and 
its corresponding hyperparameters,
the learning rate schedule 
(e.g., the initial learning rate,
the learning rate reduction multiplier, 
and how many epochs 
without improving 
the validation performance 
the algorithm waits 
before reducing the learning rate),
how many epochs without improving the validation performance
the algorithm waits before terminating 
the training process 
(i.e., early stopping), and
what 
data augmentation strategies to use and 
their corresponding hyperparameters.
The behavior of the evaluation algorithm 
with respect to the values of 
its hyperparameters is defined by the expert for the task being considered, 
so the compilation step described in Section~\ref{ssec:compilation}
for this functionality has to 
be implemented by the expert.
Nonetheless, these \emph{user hyperparameters} can be 
included in the search space and searched over in the same way as 
the architecture hyperparameters
described in Section~\ref{ssec:space_spec}.

\section{Experiments}
\label{sec:experiments}




We illustrate how our framework can be used to 
search over all hyperparameters of a model, 
i.e., both architecture and training hyperparameters, 
using only high-level insights.
We choose a search space  
of deep convolutional models
based around the 
ideas that 
depth is important, 
batch normalization helps convergence, 
and dropout is sometimes helpful.
We search over architectures and 
evaluate our models on CIFAR-10~\cite{krizhevsky2009learning}.


The training hyperparameters that we consider are
whether to use ADAM or SGD with momentum,
the initial learning rate, 
the learning rate reduction multiplier, and
the rate reduction patience, i.e., 
how many epochs without improvement to wait before reducing the 
current learning rate. 
We use standard data augmentation techniques:  
we zero pad the CIFAR-10 images 
to size $40 \times 40 \times 3$, 
randomly crop a $32 \times 32 \time 3$ portion, 
and flip horizontally at random.
We could search over these too if desired.

We compare the search algorithms described in 
Section~\ref{sec:model_search} in terms of the best model found,
according to validation performance, as a function of the 
number of evaluations. 
We run each algorithm 
$5$ times, for $64$ model evaluations each time. 
All models were trained for $30$ minutes on GeForce GTX 970
GPUs in machines with similar specifications.

On the leftmost plot of Figure~\ref{fig:searcher_comp}, 
we see that all search algorithms find 
performant solutions (around $89\%$ accuracy) after $64$ evaluations.
On the center plot of Figure~\ref{fig:searcher_comp}, we see that  
for fewer than $6$ evaluations there is considerable 
variance between the different algorithms; the more sophisticated 
model search algorithms 
are not able to outperform random search with so few evaluations. 
On the rightmost plot in Figure~\ref{fig:searcher_comp}, we see that 
both SMBO and MCTS with bisection eventually outperform random search; 
MCTS with bisection starts outperforming random search around $32$ evaluations, 
while for SMBO, it happens around $16$ evaluations.

\begin{figure}[tp]
    \centering
    \includegraphics[width=0.9\columnwidth]{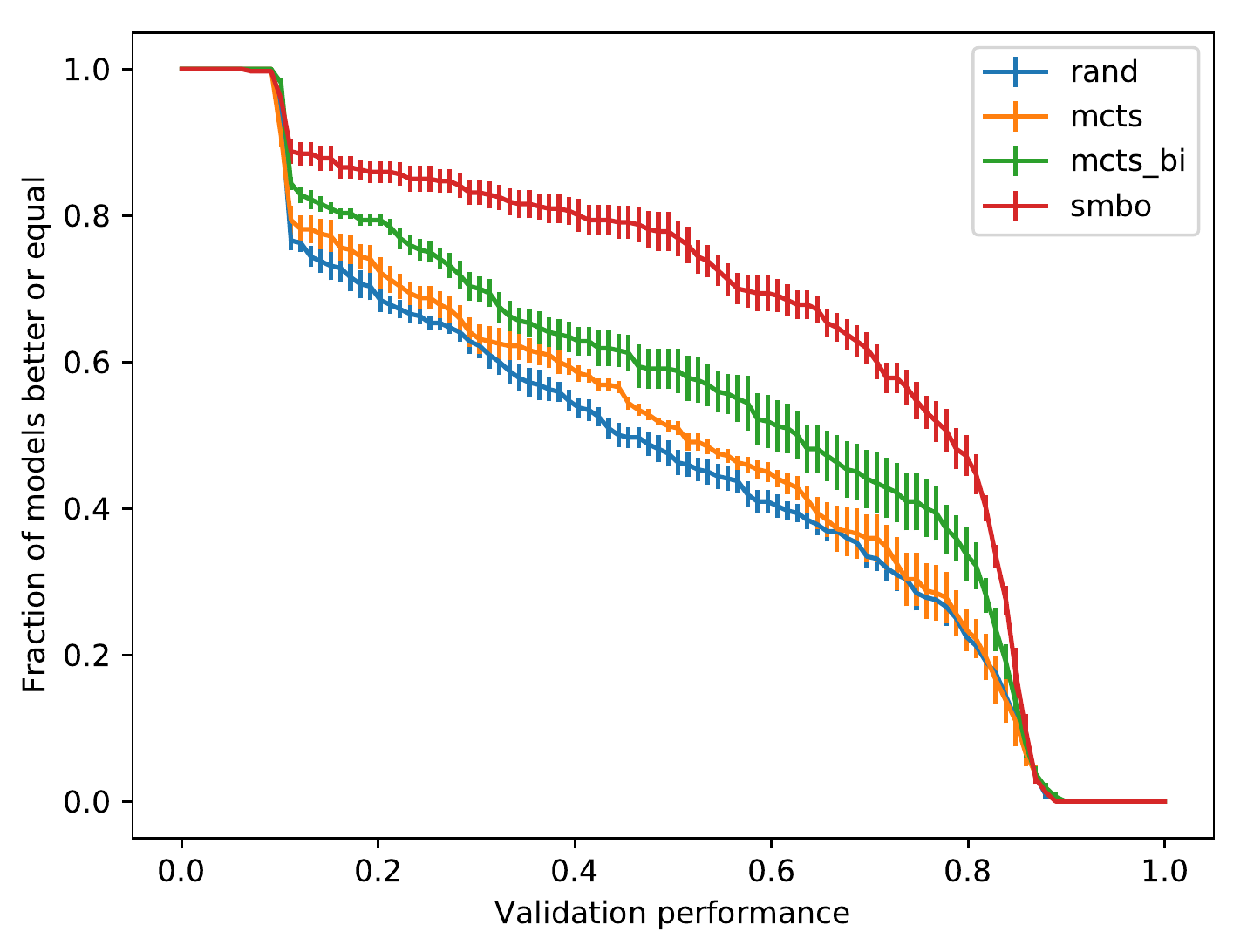}
    \caption{Percentage of models above a given validation threshold performance.
        MCTS with bisection and SMBO outperform random search.
        when compared to random search. 
        The error bars have size equal to the standard error.}
    \label{fig:quants}
\end{figure}

Surprisingly, MCTS without restructuring 
does not outperform random search. 
We think that this is because there are too many possible values for the
first few hyperparameters in the tree, so MCTS 
will not be able to 
identify and focus on high-performance regions of the search space 
within the number of evaluations available.
MCTS with bisection and SMBO do not suffer from these problems, and therefore
can identify and focus on 
high performance regions of the search space earlier.
In addition to achieving a higher top accuracy,
MCTS with bisection and 
SMBO evaluate a larger fraction of high-performance models when compared to 
random search, which can be seen in 
Figure~\ref{fig:quants}.





There is a significant number of models that
fail to train 
for the particular search space 
that we considered, i.e., 
the classification performance in the validation set 
fails to be better than random.
This is seen in 
Figure~\ref{fig:quants} by the large drop in the fraction of models 
that outperform $10 \%$ accuracy.
This observation is in accordance with the knowledge that very 
deep convolutional architectures are hard to train.
Additionally, 
in Figure~\ref{fig:quants}
we see that models 
have large variability in terms of 
the performance that they achieve in $30$ minutes of 
training. 
This is a desirable aspect of our model search specification 
language as it shows that it can represent  
expressive search spaces with models achieving 
significantly different performances.


\section{Conclusion}
\label{sec:conclusion}

We described
a framework for automatically 
designing and training deep models.
    This framework consists of three fundamental components: 
the model search space specification language, 
the model search algorithm, and 
the model evaluation algorithm.    
    The model search space specification language is composable, modular, 
and extensible, and allows us to easily define expressive search spaces 
over architectures.
The model evaluation algorithm determines how to compute a score for a model  
in the search space.
Models can be automatically compiled to their corresponding computational graphs.
Using the model search space specification language and the model evaluation algorithm, 
we can introduce model search algorithms for exploring the 
search space.
Using our framework, it is possible to do random search 
over interesting spaces of architectures
without much effort from the expert.
We also described more complex model search algorithms, such as 
MCTS, MCTS with tree restructuring, and SMBO.
We present experiments on CIFAR-10 comparing 
different model search algorithms and 
show that MCTS with tree restructuring and SMBO outperform 
random search.
Code for our framework and experiments 
has been made publicly available.
We hope that this paper will lead to more work and better tools for 
automatic architecture search.


\section*{Acknowledgements}

The authors would like to thank Matt Gormley, Matt Barnes, Christoph Dann, 
and Kirthevasan Kandasamy
for helpful discussions and detailed comments on 
early versions of the manuscript.

{\small
\bibliographystyle{ieee}
\bibliography{egbib_clean}
}

\clearpage
\appendix

\section{Detailed Experimental Setup}
\label{sec:detailed_exps}

In 
Section~\ref{sec:experiments}, 
we considered a search space of 
deep convolutional models having
structural hyperparameters 
for 
the depth of the network,
whether to apply batch normalization before or after 
ReLU, and 
whether to use dropout;
hyperparameters for the 
number and size of the convolutional filters; 
training hyperparameters for the learning rate schedule.
    We show in 
Figure~\ref{fig:ss_experiments_lisp} 
the LISP-like pseudocode for
the search space considered in Section~\ref{sec:experiments},
and in Figure~\ref{fig:ss_experiments_python}
the corresponding runnable Python implementation in our framework.

\begin{figure}[htp]
\raggedleft{ 
{\tiny
\begin{verbatim}
MH = (UserHyperparams 
        { 'optimizer_type' : ['adam', 'sgd'],
          'learning_rate_init' : logspace(10^-2, 10^-7, 32),
          'rate_mult' : logspace(10-2, 0.9, 8), 
          'rate_patience' : [4, 8, 12, 16, 20, 24, 28, 32],
          'stop_patience' : [64],
          'learning_rate_min' : [10^-9] })

M1 = (Conv2D [48, 64, 80, 96, 112, 128] [3, 5, 7] [2])

M2 = (RepeatTied 
        (Concat
            (Conv2D [48, 64, 80, 96, 112, 128] [3, 5] [1])  
            (MaybeSwap BatchNormalization ReLU)
            (Optional (Dropout [0.5, 0.9])))
        [1, 2, 4, 8, 16, 32])

M = (Concat MH M1 M2 M1 M2 (Affine [10])) 
\end{verbatim} } } 
\caption{Specification of the model search space used 
in Section~\ref{sec:experiments}
in LISP-like pseudocode. 
See Figure~\ref{fig:ss_experiments_python}
for the corresponding runnable Python code. }
\label{fig:ss_experiments_lisp}
\end{figure}

\begin{figure}[htp]
\raggedleft{ 
{\tiny
\begin{verbatim}
MH = UserHyperparams(['optimizer_type',
                      'learning_rate_init',
                      'rate_mult',
                      'rate_patience', 
                      'stop_patience', 
                      'learning_rate_min' ],
                      [['adam', 'sgd_mom'], 
                      list( np.logspace(-2, -7, num=32) ), 
                      list( np.logspace(-2, np.log10(0.9), num=8) ),
                      range(4, 33, 4), 
                      [64], 
                      [1e-9] ])

conv_initers = [ kaiming2015delving_initializer_conv(1.0) ]
aff_initers = [ xavier_initializer_affine( 1.0 )]

def Module_fn(filter_ns, filter_ls, keep_ps, repeat_ns):
    b = RepeatTied(
            Concat([
                Conv2D(filter_ns, filter_ls, [1], ["SAME"], conv_initers),
                MaybeSwap_fn( ReLU(), BatchNormalization() ),
                Optional_fn( Dropout(keep_ps) )
        ]), repeat_ns)
    return b

filter_nums = range(48, 129, 16)
repeat_nums = [2 ** i for i in xrange(6)]
mult_fn = lambda ls, alpha: list(alpha * np.array(ls))

M = Concat([MH,
            Conv2D(filter_nums, [3, 5, 7], [2], ["SAME"], conv_initers), 
            Module_fn(filter_nums, [3, 5], [0.5, 0.9], repeat_nums),
            Conv2D(filter_nums, [3, 5, 7], [2], ["SAME"], conv_initers),
            Module_fn(mult_fn(filter_nums, 2), [3, 5], [0.5, 0.9], repeat_nums),
            Affine([num_classes], aff_initers) ])
\end{verbatim} } }  
\caption{Runnable specification of the model search space used in 
Section~\ref{sec:experiments} in our Python implementation of the framework. 
See Figure~\ref{fig:ss_experiments_lisp} for the specification of the 
same search space in the LISP-like pseudocode 
used throughout this paper.}
\label{fig:ss_experiments_python}
\end{figure}

In Figure~\ref{fig:ss_experiments_lisp} 
and Figure~\ref{fig:ss_experiments_python},
to include training hyperparameters in the search space, 
we concatenate
the module that encapsulates the training hyperparameters 
(the module assigned to \texttt{MH}) 
and the modules that encapsulate the 
remaining model hyperparameters 
(the modules other than \texttt{MH} 
in the declaration of \texttt{M}).

The Python specification 
of the model search space
in Figure~\ref{fig:ss_experiments_python} is remarkably close in both 
semantics and length 
to the LISP-like pseudocode in Figure~\ref{fig:ss_experiments_lisp}.
We omit some hyperparameters in Figure~\ref{fig:ss_experiments_lisp}
because we did not consider multiple values for them, e.g., 
for \texttt{Conv2D} modules, we always used 
same size padding
and the
initialization scheme described in He \etal~\cite{he2015delving}.

Our implementation has code modularity and reusability benefits.
For example, 
we can define an auxiliary function to instantiate modules and then 
use it in the instantiation of the module for the complete search space.
This is illustrated in Figure~\ref{fig:ss_experiments_python} 
with the definition of \texttt{Module\_fn}
and its use in the declaration of \texttt{M}. 

\section{List of Modules}
\label{sec:list_modules}

We provide a brief description of 
a representative subset of 
the types of basic and composite modules that
we have implemented in our framework.

\subsection{Basic Modules}


Basic modules take no  
other modules when instantiated,
having only local hyperparameters
and parameters.

\begin{itemize}
    \item \texttt{Affine}: 
        Dense affine transformation. 
        Hyperparameters: 
            number of the hidden units and
            initialization scheme of the parameters.
        Parameters: 
            dense matrix and
            bias vector.
    \item \texttt{ReLU}: 
        ReLU nonlinearity. 
        Hyperparameters: none.
        Parameters: none.  
    \item \texttt{Dropout}: 
        Dropout. 
        Hyperparameter: 
            dropout probability.
        Parameters: 
            none.
    \item \texttt{Conv2D}: 
        Two-dimensional convolution. 
        Hyperparameters: 
            number of filters,
            size of the filters,
            stride,
            padding scheme, and
            initialization scheme of the parameters.
        Parameters: 
            convolutional filters and
            bias vector.
    \item \texttt{MaxPooling2D}:
        Two-dimensional max pooling.
        Hyperparameters: 
            size of the filters,
            stride, and
            padding scheme.
        Parameters:
            none.
    \item \texttt{BatchNormalization}:
        Batch normalization.
        Hyperparameters:
            none.
        Parameters:
            translation coefficients and
            scaling coefficients.
    \item \texttt{UserHyperparams}:
        User-defined hyperparameters.
        Hyperparameters:
            hyperparameters determined by the user expert.
        Parameters:
            none.
    \item \texttt{Empty}:
        Identity.
        Hyperparameters:
            none.
        Parameters:
            none.
\end{itemize}

\subsection{Composite Modules}

Composite modules take other modules as arguments 
when instantiated, 
which we will call \emph{submodules}.
The behavior of a composite module depends 
on its submodules.
The hyperparameters 
which a 
composite module has to specify 
depend on the values of the hyperparameters of the composite module
and the hyperparameters of the submodules;
e.g., \texttt{Or} takes a list of submodules 
but it only has to specify 
the hyperparameters of the 
submodule that it ends up choosing.
A composite module  
is responsible for specifying its submodules, 
which is done through calls to 
the module interfaces of the submodules.

\begin{itemize}
    \item \texttt{Concat}: 
        Takes a list of submodules and connects them 
            in series.
        Hyperparameters:
            hyperparameters of the submodules.
        Parameters:
            parameters of the submodules.
    \item \texttt{Or}: 
        Chooses one of its submodules to use.
        Hyperparameters:
            which submodule to use and
            hyperparameters of the submodule chosen.
        Parameters:
            parameters of the submodule chosen.
    \item \texttt{Repeat}: 
        Repeats a submodule some number of times,
        connecting the repetitions in series; 
        values for the hyperparameters of the repetitions
        are chosen independently.
        Hyperparameters:
            number of times to repeat the submodule 
            and
            hyperparameters of the repetitions of the submodule.
        Parameters:
            parameters of the repetitions of the submodule.
    \item \texttt{RepeatTied}: 
        Same as \texttt{Repeat}, but values for the hyperparameters of the 
        submodule are chosen once and used for all the submodule repetitions.
        Hyperparameters:
            the number of times to repeat the submodule and 
            hyperparameters of the submodule.
        Parameters:
           parameters of the repetitions of the submodule.
    \item \texttt{Optional}: 
        Takes a submodule and chooses whether to use it or not.
        Hyperparameters:
            whether to include the submodule or not and, 
            if included, 
            hyperparameters of the submodule.
        Parameters:
            if included, parameters of the submodule.
    \item \texttt{Residual}:
        Takes a submodule and implements a skip connection adding 
        the input and output;
        if the input and output have different dimensions, 
        they are padded to make addition possible.
        Hyperparameters:
            hyperparameters of the submodule.
        Parameters:
            parameters of the submodule.
    \item \texttt{MaybeSwap}: 
        Takes two submodules and connects them in series, 
        choosing which submodule comes first.
        Hyperparameters:
            which of the submodules comes first
            and hyperparameters of the submodules.
        Parameters:
            parameters of the submodules.        
\end{itemize}

\section{Module Interface}

We describe the module interface as we implemented it in 
Python. To implement a new type of module, one only 
needs to implement the module interface.

\begin{figure}[htp]
{\raggedleft 
{\scriptsize
\begin{verbatim}
class Module(object):
    def initialize(self, in_d, scope)
    def get_outdim(self)
    def is_specified(self)
    def get_choices(self)
    def choose(self, choice_i)
    def compile(self, in_x, train_feed, eval_feed)
\end{verbatim} } }
\caption{Module interface used by all modules irrespective if they are 
basic or composite. To implement a new type of module, the human expert only 
needs to implement the module interface.}
\label{fig:interface} 
\end{figure}

\begin{itemize}
    \item \texttt{initialize}: 
        Tells a module its input dimensionality.
        A composite module is responsible 
        for initializing the submodules 
        that it uses.

    \item \texttt{get\_outdim}:
        Once a module is fully specified, we can determine its output dimensionality
        by calling \texttt{get\_outdim}. 
        The output dimensionality is a function of the input dimensionality 
        (which is determined when \texttt{initialize} is called)
        and the values of the hyperparameters chosen.
    \item \texttt{is\_specified}:
        Tests whether a module is fully specified.
        If a module is fully specified, \texttt{outdim} and \texttt{compile}
        may be called.
    \item \texttt{get\_choices}:
        Returns a list of the possible values for the hyperparameter 
        currently being specified.
    \item \texttt{choose}:
        Chooses one of the possible values for the hyperparameter 
        being specified. 
        The module assigns the chosen value to that hyperparameter and either
        transitions to the next hyperparameter to 
        specify or becomes fully specified. 
        The module maintains 
        internally the state of its search process.



    \item \texttt{compile}:
        Creates the computational graph of the model in a deep learning 
        model specification language, such as Tensorflow or PyTorch.
        For composite modules, compilation can be performed recursively, 
        through calls to the \texttt{compile} functions of its submodules.
\end{itemize}

Composite modules rely on calls to the module interfaces of its submodules 
to implement their own module interfaces. For example, \texttt{Concat} needs
to call \texttt{out\_dim} for the last submodule of the series connection 
to determine its own output dimensionality,
and needs to call \texttt{choose} on the submodules to specify itself.
One of the design choices that make the language modular 
is the fact that a \emph{composite module can 
implement its own module interface through calls to the module interfaces of 
its submodules}.
All information about the specification of a module is local 
to itself or kept within its submodules.

\section{Beyond Single-Input Single-Output Modules}
\label{sec:beyond_siso}

We can define new modules with 
complex signal paths as long as their existence is encapsulated, i.e.,  
a module may have 
many signal paths
as long they
fork from a single input and 
merge to a single output, 
as illustrated in 
Figure~\ref{fig:complex_module}. 

\tikzset{
    module/.style = {draw, fill=white, rectangle, minimum height=2em, minimum width=3em},
    tmp/.style  = {coordinate}, 
    input/.style = {coordinate},
    output/.style= {coordinate},
    sum/.style= {draw, fill=white, circle},
}

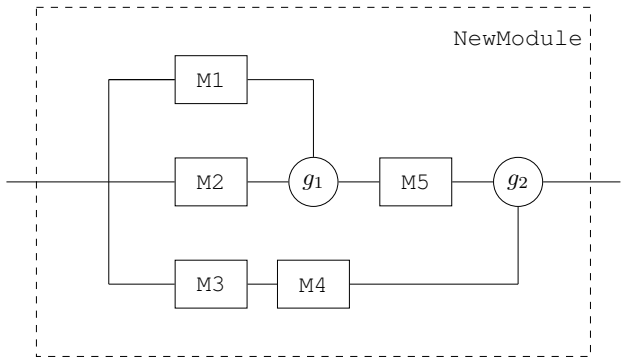
\begin{figure}[htp]
\centering
    \resizebox{\columnwidth}{!}{
        \begin{tikzpicture}[auto, node distance=1.5cm,>=latex', scale=0.50]
            \node [module] (b2) {\texttt{M2}};
            \node [module, above of=b2] (b1) {\texttt{M1}};
            \node [module, below of=b2] (b3) {\texttt{M3}};
            \node [input, left of=b1] (in_m1) {};
            \node [input, left of=b2] (in_m2) {};
            \node [input, left of=b3] (in_m3) {};
            \node [input, left of=in_m2] (in_m) {};
            \draw [-] (in_m) -- (in_m2);
            \draw [-] (in_m1) -- (in_m2);
            \draw [-] (in_m2) -- (in_m3);
            \draw [-] (in_m1) -- (b1);   
            \draw [-] (in_m1) -- (b1);   
            \draw [-] (in_m2) -- (b2);   
            \draw [-] (in_m3) -- (b3);   
            \node [sum, right of=b2] (g1) {$g _1$};
            \node [tmp, right of=b1] (tmp1_g1) {};
            \draw [-] (b2) -- (g1);
            \draw [-] (b1) -- (tmp1_g1);
            \draw [-] (tmp1_g1) -- (g1);    
            \node [module, right of=b3] (b4) {\texttt{M4}};
            \draw [-] (b3) -- (b4);
            \node [module, right of=g1] (b5) {\texttt{M5}};
            \draw [-] (g1) -- (b5);
            \node [sum, right of=b5] (g2) {$g _2$};
            \node [tmp, below of=g2] (tmp4_g2) {};
            \draw [-] (b5) -- (g2);
            \draw [-] (b4) -- (tmp4_g2);
            \draw [-] (tmp4_g2) -- (g2);
            \node [output, right of=g2] (out_b) {};
            \draw [-] (g2) -- (out_b) {};
            \node [tmp, above left of=in_m1] (tmp_lu) {};
            \node [tmp, below left of=in_m3] (tmp_ld) {};
            \node [tmp, above of=g2] (tmp_ru_aux) {};
            \node [tmp, above right of=tmp_ru_aux] (tmp_ru) {};
            \node [tmp, below right of=tmp4_g2] (tmp_rd) {};
            \draw [dashed] (tmp_lu) -- (tmp_ru) -- (tmp_rd) -- (tmp_ld) -- (tmp_lu);
            \node [tmp, below of=tmp_ru] (tmp_anot) {};
            \node [above left of=tmp_anot] {\texttt{NewModule}};
        \end{tikzpicture}
    }
    \caption{A module with many signal paths from input to output. 
        To implement a module, the human expert 
        only needs to implement its 
        module interface. 
        \texttt{M1}, \texttt{M2}, \texttt{M3}, and \texttt{M4} are 
        arbitrary  
        single-input single-output modules; 
        $g _1$ and $g _2$ are arbitrary transformations 
        that may have additional hyperparameters. 
        The hyperparameters of $g _1$ and $g _2$  
        can be managed internally by \texttt{NewModule}. }
    \label{fig:complex_module}
\end{figure}

In Figure~\ref{fig:complex_module} there is a single 
input fed into
\texttt{M1}, \texttt{M2}, and \texttt{M3}. 
\texttt{M1}, \texttt{M2}, \texttt{M3}, \texttt{M4}, \texttt{M5} 
are arbitrary single-input single-output 
submodules of \texttt{NewModule}.
The module interface of \texttt{NewModule} 
can be implemented
using the module interfaces of its submodules.
Instantiating a module of type \texttt{NewModule} requires submodules 
for \texttt{M1}, \texttt{M2}, \texttt{M3}, \texttt{M4}, and \texttt{M5},
and potentially lists of possible values 
for the hyperparameters of $g _1$ and $g _2$.
A residual module which chooses what
type of merging function to apply, e.g., additive or multiplicative,
is an example of a module with hyperparameters for the merging functions 

A module of the type \texttt{NewModule} is fully specified after 
we choose values for all the hyperparameters of  
\texttt{M1}, \texttt{M2}, \texttt{M3}, \texttt{M4}, \texttt{M5}, $g_1$,
and $g _2$. 
Testing if \texttt{M1}, \texttt{M2}, \texttt{M3}, \texttt{M4}, 
and \texttt{M5} are fully specified 
can be done by calling 
\texttt{is\_specified} 
on the corresponding submodule.
    
    The output dimensionality of \texttt{NewModule} can be computed
as a function of 
the 
values of the hyperparameters of $g _2$
and 
the output dimensionality of \texttt{M5} 
and \texttt{M4}, which 
can be obtained 
by calling \texttt{get\_outdim}.
Similarly, for \texttt{get\_choices} 
we have to keep track of which 
hyperparameter we are specifying, 
which can either come from 
\texttt{M1}, \texttt{M2}, \texttt{M3}, \texttt{M4}, and \texttt{M5}, 
or from 
$g _1$ and $g _2$. 
If we are choosing values for an hyperparameter in 
\texttt{M1}, \texttt{M2}, \texttt{M3}, \texttt{M4}, and \texttt{M5}
we can call \texttt{get\_choices} and \texttt{choose} on that submodule, 
while for the hyperparameters of $g _1$ and $g _2$ 
we have to keep track of the state in \texttt{NewModule}.
\texttt{compile} is similar 
in the sense that it is
implemented using calls
to the \texttt{compile} functionality of 
the submodules.

\end{document}